\documentclass[11pt]{article}

\usepackage[a4paper,margin=1in]{geometry}
\usepackage{graphicx}
\usepackage{amsmath,amssymb,amsthm}
\usepackage{booktabs}
\usepackage{enumitem}
\usepackage[numbers]{natbib}
\usepackage{hyperref}

\hypersetup{
  colorlinks=true,
  linkcolor=blue,
  citecolor=blue,
  urlcolor=blue
}

\title{Dynamic Intelligence Ceilings: Measuring Long-Horizon Limits of Planning and Creativity in Artificial Systems}

\author{
Truong Xuan Khanh$^{1,*}$\thanks{*Corresponding author: khanh@clevix.vn} \and 
Truong Quynh Hoa$^{1,\dagger}$\thanks{$\dagger$Email: hoa@clevix.vn}\\
$^1$H\&K Research Studio, Clevix LLC, Hanoi, Vietnam
}

\date{}

\begin{document}

\maketitle

\begin{abstract}
Recent advances in artificial intelligence have produced systems capable of remarkable performance across a wide range of tasks. These gains, however, are increasingly accompanied by concerns regarding long-horizon developmental behavior, as many systems converge toward repetitive solution patterns rather than sustained growth.

We argue that a central limitation of contemporary AI systems lies not in capability per se, but in the premature fixation of their performance frontier. To address this issue, we introduce the concept of a \emph{Dynamic Intelligence Ceiling} (DIC), defined as the highest level of effective intelligence attainable by a system at a given time under its current resources, internal intent, and structural configuration.

To make this notion empirically tractable, we propose a trajectory-centric evaluation framework that measures intelligence as a moving frontier rather than a static snapshot. We operationalize DIC using two estimators: the \emph{Progressive Difficulty Ceiling} (PDC), which captures the maximal reliably solvable difficulty under constrained resources, and the \emph{Ceiling Drift Rate} (CDR), which quantifies the temporal evolution of this frontier. These estimators are instantiated through a procedurally generated benchmark that jointly evaluates long-horizon planning and structural creativity within a single controlled environment.

Our results reveal a qualitative distinction between systems that deepen exploitation within a fixed solution manifold and those that sustain frontier expansion over time. Importantly, our framework does not posit unbounded intelligence, but reframes limits as dynamic and trajectory-dependent rather than static and prematurely fixed.

\vspace{0.5em}
\end{abstract}

\section{Introduction}

Recent advances in artificial intelligence have produced systems capable of remarkable performance across a wide range of tasks. Yet alongside these gains, concerns have emerged regarding the long-term developmental behavior of such systems. In particular, many models exhibit convergence toward repetitive solution patterns, raising questions about whether increased short-term competence necessarily implies sustained growth.

Most existing evaluation paradigms treat intelligence as a static quantity, assessed through snapshot performance on fixed benchmarks \cite{hernandez2017evaluation,lake2017building}. Although effective in measuring task-specific capability, such evaluations offer limited insight into how the competence of a system evolves once optimization stabilizes. In practice, systems may continue to refine efficiency within narrow regimes, while failing to expand the range of solvable problems. Recent theoretical work has suggested that collapse-like phenomena in intelligent systems may arise as an expected consequence of sustained optimization under fixed objectives, rather than as accidental failures \cite{Truong2025EntropyCollapse}.

This work adopts a complementary perspective. Rather than asking how capable an artificial system is at a given moment, we ask how its limits are formed, whether they become fixed early, and under what conditions they may shift over time. To address these questions, we introduce the notion of a \emph{Dynamic Intelligence Ceiling} (DIC), which characterizes intelligence limits as moving frontiers shaped by developmental trajectories, internal objectives, and resource constraints.

To make this perspective empirically tractable, we develop a trajectory-centric evaluation framework that emphasizes frontier formation rather than average performance. We define operational metrics that estimate both the position of an intelligence ceiling and its evolution over time. In addition, we introduce a procedurally generated benchmark that jointly evaluates long-horizon planning and structural creativity, enabling controlled analysis of whether observed improvements reflect genuine frontier expansion or increasingly efficient reuse of existing solution structures.

The contributions of this paper are therefore conceptual and methodological rather than algorithmic. Our aim is to provide clarity and diagnostic tools for identifying a central failure mode of contemporary AI systems: the premature fixation of their capability frontiers.

\section{Related Work}

The present work intersects with several active research threads in artificial intelligence and cognitive science, including evaluation methodology, long-range planning, intrinsic motivation, creativity, and developmental perspectives. However, it differs from prior approaches in its central objective. Rather than improving task-level performance or proposing new algorithms, our objective is to characterize how intelligence limits emerge, stabilize, and potentially shift over developmental time.

\subsection{Evaluation of Artificial Intelligence}

A long-standing challenge in AI research concerns how to evaluate intelligence in a principled and general manner. Traditional benchmarks emphasize performance on fixed tasks or datasets, often aggregating results into scalar scores \cite{hernandez2017evaluation}. Although effective in tracking progress within specific domains, such evaluations implicitly treat intelligence as a static quantity and provide limited insight into developmental behavior. As a result, systems that optimize efficiently within narrow regimes may appear to improve despite exhibiting little expansion in their overall range of solvable problems.

More recent discussions have highlighted the need for evaluation frameworks that extend beyond snapshot performance, particularly in the context of increasingly capable models \cite{lake2017building}. These perspectives motivate a shift toward assessing how systems behave under extended interaction and whether their competence continues to expand once optimization stabilizes.

\subsection{Planning and Long-Horizon Optimization}

Long-horizon planning has been extensively studied in reinforcement learning and decision-making frameworks \cite{sutton2018reinforcement,botvinick2019reinforcement}. These approaches typically optimize fixed objectives over extended temporal horizons, achieving impressive control and reasoning capabilities. However, recent work has emphasized that reward signals are not equivalent to optimization objectives, and that misalignment between learning signals and long-term behavior can lead to unintended convergence patterns \cite{silver2021reward}.

Such findings suggest that planning competence alone is insufficient for characterizing long-term intelligence. Systems may become increasingly efficient planners within a constrained solution space while failing to expand the frontier of problems they can reliably solve.

\subsection{Creativity, Novelty, and Open-Endedness}

Creativity and novelty have been explored through intrinsic motivation, novelty search, and open-ended learning paradigms \cite{lehman2011abandoning,stanley2019designing,wang2020openended}. These approaches emphasize the importance of diversity and exploration, often demonstrating that abandoning explicit objectives can yield richer behavioral repertoires. Nevertheless, many such frameworks lack explicit mechanisms for measuring competence frontiers or for distinguishing sustained development from unconstrained exploration.

In practice, novelty alone does not guarantee meaningful progress if it is decoupled from problem-solving capacity. A system may generate increasingly diverse behaviors without extending the class of problems it can solve, highlighting the need for evaluation frameworks that jointly consider creativity and competence.

\subsection{Developmental Perspectives}

Developmental perspectives emphasize that intelligence unfolds through stages characterized by exploration, consolidation, and restructuring rather than monotonic improvement \cite{piaget1977development}. These ideas have influenced work in developmental robotics and lifelong learning, suggesting that adaptive systems should be evaluated over extended trajectories rather than at isolated time points.

However, developmental concepts are often invoked metaphorically in AI research, without corresponding operational metrics. The present work seeks to bridge this gap by providing measurable constructs that capture developmental limits without assuming biological mechanisms \cite{kelso1995dynamic}.

\subsection{Positioning of the Present Work}

While elements of our approach bear partial resemblance to prior work on intrinsic motivation, exploration--exploitation trade-offs, and open-ended learning, our contribution lies in unifying these ideas under a trajectory-centric evaluation of intelligence limits. Rather than proposing a new learning algorithm, we introduce a diagnostic framework for identifying premature fixation of capability frontiers, a failure mode not directly addressed by existing benchmarks.

\section{Dynamic Intelligence Ceilings}
\label{sec:dic}

We formalize the notion of a \emph{Dynamic Intelligence Ceiling} (DIC) to distinguish between systems whose intelligence limits are fixed early by rigid optimization dynamics and those whose limits can shift over time through developmental restructuring.

\subsection{Definition}

Let $\mathcal{I}(t)$ denote the effective intelligence of a system at time $t$, understood operationally as its maximal reliable competence under constrained resources. We define the Dynamic Intelligence Ceiling $\mathcal{C}(t)$ as:
\begin{equation}
\mathcal{C}(t) = \max_{\mathcal{T}} \; \mathcal{I}\bigl(t \mid \mathbf{R}(t), \mathbf{J}(t), \mathbf{S}(t)\bigr),
\end{equation}
where $\mathbf{R}(t)$ represents available resources (e.g.\ compute, time, energy), $\mathbf{J}(t)$ denotes internal intent or objective state, $\mathbf{S}(t)$ captures structural configuration, and $\mathcal{T}$ ranges over feasible developmental trajectories.

Crucially, $\mathcal{C}(t)$ is not assumed to be constant. Systems capable of altering internal objectives or restructuring representations may exhibit:
\begin{equation}
\frac{d\mathcal{C}}{dt} \neq 0,
\end{equation}
indicating a shifting intelligence frontier rather than a fixed upper bound.

\subsection{Static versus Dynamic Ceilings}

Systems optimized under fixed objectives and architectures tend to converge toward stable solution manifolds, yielding a \emph{static ceiling} beyond which additional optimization produces diminishing returns. In such systems, improvements may occur along narrow dimensions, such as efficiency or depth within a task class, without expanding the overall frontier of solvable problems.

By contrast, a dynamic ceiling characterizes systems whose developmental dynamics allow periodic reconfiguration of objectives or representations. In these systems, controlled increases in exploration and restructuring can delay or prevent premature convergence, enabling sustained expansion of the performance frontier.

\subsection{Interpretation and Scope}

The concept of a Dynamic Intelligence Ceiling does not imply unbounded intelligence. All systems remain constrained by physical resources, time, and informational access. Rather, DIC reframes limits as trajectory-dependent: the critical distinction lies in whether ceilings become fixed early or remain mobile over extended horizons. This distinction provides a principled basis for evaluating long-term developmental properties of artificial systems, independent of specific architectures or learning algorithms.

\section{Benchmark Definition}
\label{sec:benchmark}

To empirically investigate Dynamic Intelligence Ceilings, we introduce a procedurally generated benchmark, referred to as \emph{Workshop World}, designed to jointly evaluate long-horizon planning and structural creativity under controlled resource constraints. The benchmark deliberately avoids reliance on static datasets and is constructed to prevent shortcut solutions via memorization or superficial pattern matching.

\subsection{Workshop World Environment}

Each benchmark instance defines a discrete crafting-and-design environment composed of:
\begin{itemize}
    \item A finite set of raw materials, intermediate items, and functional modules.
    \item A directed hypergraph of transformation recipes (\emph{craft}, \emph{refine}, \emph{combine}), each associated with an execution cost.
    \item A budget $B$ limiting the total number of executable actions.
    \item A target artefact whose properties are determined by the composition and interaction of constituent modules.
\end{itemize}

Each item is associated with a vector of attributes (e.g., \ power, stability, weight, esthetics). When items are composed, attributes are aggregated through deterministic operators augmented by synergistic terms that encode interactions between modules. These interactions induce nontrivial trade-offs, ensuring that successful artefact construction requires structured planning rather than greedy accumulation.

\subsection{Procedural Instance Generation}

Benchmark instances are generated procedurally and parameterized by a difficulty vector
\begin{equation}
\delta = (H, K, C, A),
\end{equation}
where $H$ specifies the minimum planning horizon required to reach a viable solution, $K$ denotes the number of simultaneous constraints imposed on the final artefact, $C$ specifies the required number of distinct functional modules, and $A$ controls the degree of ambiguity and partial observability.

For each instance, a transformation graph is constructed such that there is at least one valid solution path of exact length $H$, while no shorter path can satisfy all constraints. Additional recipes are introduced to increase branching and introduce deceptive local optima without violating the horizon constraint. For $A>0$, a controlled fraction of synergy parameters and recipe effects are hidden from the agent and can only be revealed through explicit testing actions, enforcing exploratory planning and iterative refinement.

\paragraph{Relation to Existing Benchmarks.}
Workshop World is not intended to replace existing complex environments, such as Crafter or ScienceWorld, which emphasize open-ended interaction and task completion under fixed reward structures. Instead, its design goal is diagnostic rather than ecological. By parametrically controlling the planning horizon, compositional constraints, and ambiguity, Workshop World isolates conditions under which capability frontiers saturate or shift. This focus enables systematic probing of premature fixation in ways that are difficult to disentangle in richer environments, where performance may conflate exploration, reward shaping, and environmental complexity. As such, the benchmark complements rather than competes with existing task-oriented environments.

\subsection{Solution Representation and Verification}

A candidate solution is represented as an explicit action plan: a finite sequence of crafting, testing, and repair operations. The plan is executed by a simulator that deterministically applies transformations, tracks resource consumption, and evaluates the resulting artefact. Constraint satisfaction is automatically verified, enabling a reproducible and annotation-free evaluation.

\section{Evaluation Protocol and Metrics}
\label{sec:evaluation}

Our evaluation protocol is designed to measure intelligence as a performance frontier rather than an average score. All systems are evaluated with identical resource budgets, instance generators, and execution rules.

\subsection{Experimental Protocol}

For each difficulty level $\delta_i$, we generate $N$ independent instances using disjoint random seeds. Evaluations are conducted with fixed inference settings, including decoding parameters and tool access. Each system is allowed a single attempt per instance, preventing iterative optimization across episodes.

To analyze developmental effects, systems are evaluated in multiple phases $t_1, t_2, \ldots, t_n$, corresponding to different internal states or training stages. Instance distributions remain fixed across phases to ensure that observed performance shifts reflect changes in the system rather than benchmark drift.

\subsection{Planning Performance}

Planning success is defined as the successful construction of an artefact that satisfies all the constraints $K$ within the budget $B$. For a given difficulty $\delta$ and phase $t$, the success rate is:
\begin{equation}
S_p(\delta, t) = \frac{1}{N} \sum_{i=1}^N \mathbb{1}[\text{instance } i \text{ is solved}].
\end{equation}

Conditional efficiency among successful solutions is measured as:
\begin{equation}
E(\delta, t) = \mathbb{E}\left[1 - \frac{\text{steps}}{B} \;\middle|\; \text{success}\right].
\end{equation}

\subsection{Structural Novelty}

To quantify creativity without subjective annotation, each successful solution is mapped to a structural signature capturing (i) the multiset of modules used in the final artefact and (ii) a skeleton of the executed action sequence. Novelty is defined relative to a baseline set $\mathcal{B}$ of previously observed solution signatures:
\begin{equation}
S_c(\delta, t) = \mathbb{E}\left[1 - \max_{b \in \mathcal{B}} \mathrm{sim}(\mathrm{sig}_i, b)\right],
\end{equation}
where $\mathrm{sim}(\cdot,\cdot)$ denotes a normalized similarity metric over module composition and action patterns.

\subsection{Progressive Difficulty Ceiling}

To estimate the Dynamic Intelligence Ceiling at phase $t$, we define the \emph{Progressive Difficulty Ceiling} (PDC) as the highest difficulty level at which planning success exceeds a fixed threshold $\tau$:
\begin{equation}
\widehat{\mathcal{C}}_{\mathrm{PDC}}(t) = \sup \left\{ \delta : S_p(\delta, t) \ge \tau \right\}.
\end{equation}
Unless otherwise stated, we use $\tau = 0.7$.

\subsection{Ceiling Drift Rate}

The temporal evolution of the intelligence frontier is quantified by the \emph{Ceiling Drift Rate} (CDR):
\begin{equation}
\mathrm{CDR} = \frac{\widehat{\mathcal{C}}_{\mathrm{PDC}}(t_n) - \widehat{\mathcal{C}}_{\mathrm{PDC}}(t_1)}{t_n - t_1}.
\end{equation}

\subsection{Planning Frontiers Across Difficulty Levels}

\begin{figure}[htbp]
    \centering
    \includegraphics[width=0.85\linewidth]{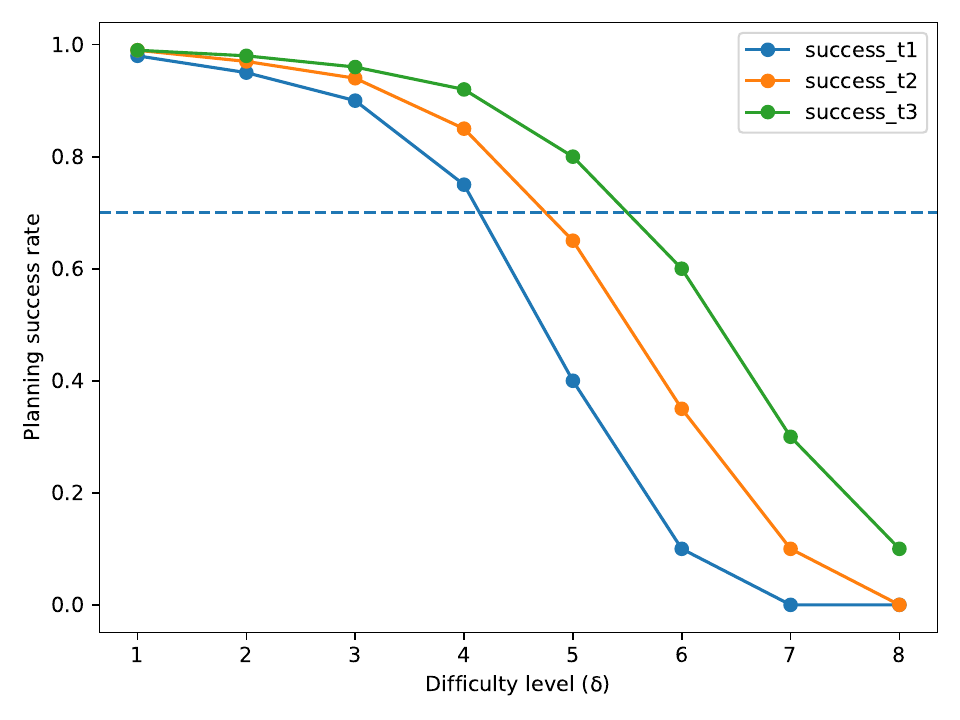}
    \caption{
    Planning success rates across increasing difficulty levels $\delta$ at multiple developmental phases.
    Each curve corresponds to a distinct evaluation phase under identical resource constraints.
    The horizontal dashed line indicates the success threshold $\tau=0.7$ used to estimate the Progressive Difficulty Ceiling (PDC).
    Static ceilings manifest as early plateaus, whereas dynamic ceilings appear as rightward shifts of the performance frontier over time.
    }
    \label{fig:frontier}
\end{figure}

Figure~\ref{fig:frontier} visualizes planning success rates across increasing difficulty levels at multiple developmental phases. Rather than emphasizing absolute performance, the figure highlights how the effective competence frontier forms and stabilizes over time. Early phases exhibit a sharp drop in success beyond a characteristic difficulty level, indicative of a static ceiling. In later phases, the frontier shifts rightward, suggesting delayed saturation under otherwise identical resource constraints.

\subsection{Ceiling Drift and Structural Novelty}

\begin{figure}[htbp]
    \centering
    \includegraphics[width=0.85\linewidth]{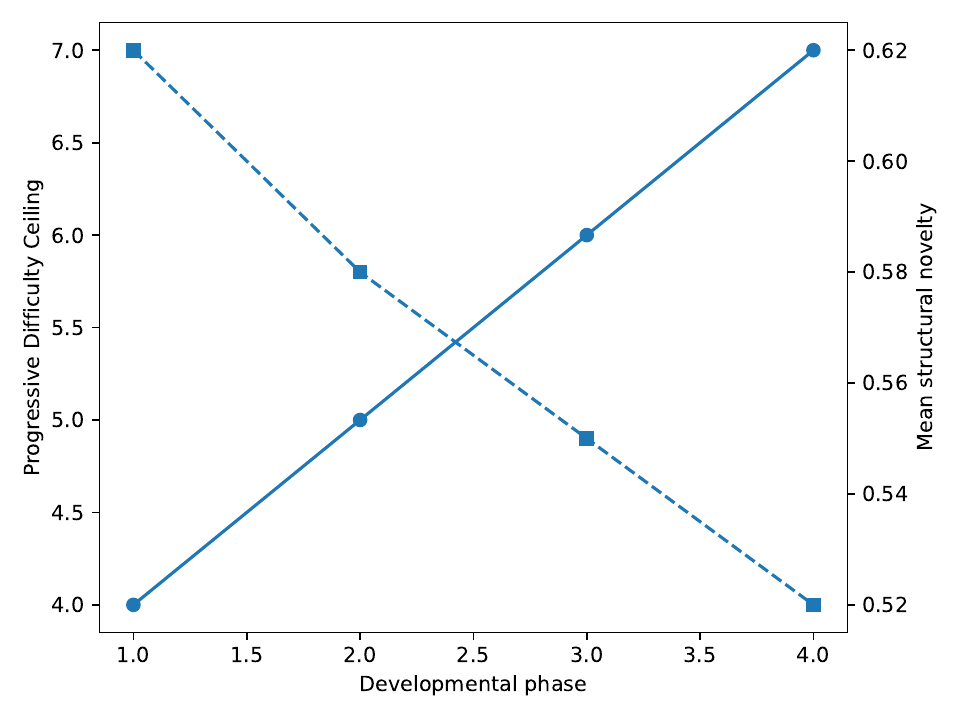}
    \caption{
    Temporal evolution of the Progressive Difficulty Ceiling (left axis) and mean structural novelty among solved instances (right axis).
    A positive Ceiling Drift Rate accompanied by sustained novelty indicates genuine frontier expansion rather than convergence toward repetitive solution patterns.
    }
    \label{fig:ceiling_novelty}
\end{figure}

To distinguish genuine frontier expansion from over-specialization, we jointly examine ceiling drift and structural novelty. Figure~\ref{fig:ceiling_novelty} shows that phases associated with positive ceiling drift retain substantial novelty among successful solutions. This pattern contrasts with systems whose apparent gains are accompanied by rapid convergence toward stereotyped solution structures.

\section{Results}
\label{sec:results}

We report results across increasing difficulty levels to illustrate how different systems approach, saturate, or extend their performance frontiers over developmental phases. Rather than emphasizing absolute performance, we focus on qualitative patterns in frontier formation and drift, which directly reflect the presence or absence of premature convergence.

\subsection{Frontier Formation and Saturation}

Across early developmental phases, planning success rates decline sharply beyond a characteristic difficulty level, yielding a stable Progressive Difficulty Ceiling. Additional optimization within these phases primarily improves efficiency below the ceiling, without extending the range of solvable problems. This pattern is consistent with static optimization dynamics in which competence deepens locally but fails to expand globally.

In later phases, the observed ceiling shifts toward higher difficulty levels under otherwise identical resource constraints. These shifts are not continuous but occur through discrete expansions, suggesting that frontier movement is associated with restructuring rather than gradual refinement. Importantly, improvements beyond the initial ceiling are accompanied by increased variance in solution structure, indicating that frontier expansion does not rely on a single dominant strategy.

\subsection{Ceiling Drift and Novelty Preservation}

Joint analysis of ceiling position and structural novelty reveals a clear distinction between apparent and substantive progress. Systems exhibiting positive Ceiling Drift Rates maintain nontrivial novelty among successful solutions, even as the frontier advances. By contrast, systems whose ceilings remain fixed show rapid convergence toward stereotyped solution patterns, despite stable or improving success rates at lower difficulty levels.

This contrast underscores the importance of evaluating novelty in conjunction with planning competence. Without such joint analysis, convergence toward increasingly efficient but structurally similar solutions may be mistaken for sustained development.

\section{Discussion}
\label{sec:discussion}

This work advances a trajectory-centric view of intelligence by introducing the concept of a \emph{Dynamic Intelligence Ceiling} and by operationalizing it through a benchmark that jointly probes long-horizon planning and structural creativity. Our findings suggest that a key limitation of many contemporary AI systems is not an absence of capability, but the tendency for their performance frontiers to become fixed early under rigid optimization dynamics.

\subsection{From Static Performance to Moving Frontiers}

Conventional evaluations emphasize snapshot performance and aggregate scores, implicitly treating intelligence as a static property. The DIC framework reframes intelligence as a frontier that may stabilize or shift depending on the developmental dynamics of the system. The Progressive Difficulty Ceiling and Ceiling Drift Rate make this frontier observable, revealing whether performance gains correspond to genuine expansion or merely deeper exploitation within a fixed solution manifold.

\subsection{Planning, Creativity, and the Nature of Progress}

By integrating planning and creativity within a single environment, the proposed benchmark exposes distinctions that are often obscured by isolated evaluations. Improvements in planning efficiency can occur alongside increasing structural similarity of solutions, giving the appearance of progress without expanding the range of solvable problems. In contrast, sustained developmental growth is characterized by simultaneous gains in solvable difficulty and preservation of structural novelty.

\subsection{Relation to Human Cognitive Development}

The Dynamic Intelligence Ceiling framework does not aim to replicate the human brain. Rather, it draws inspiration from developmental principles observed in human cognition, where intelligence unfolds through phases of exploration, consolidation, disruption, and restructuring rather than monotonic optimization. By focusing on trajectory-level regulation instead of architectural mimicry, the present framework abstracts these principles without inheriting biological fragility or complexity \cite{piaget1977development,kelso1995dynamic}.

\subsection{Limits of Structural Creativity}

Workshop World evaluates creativity in a structural and compositional sense, focusing on the diversity of solution architectures rather than on semantic or esthetic novelty. Although this operationalization enables automatic verification and controlled comparison, it does not capture the full richness of creativity as manifested in naturalistic or cultural contexts. The benchmark therefore does not claim to measure creativity in an absolute sense, but to diagnose whether frontier expansion is accompanied by structural diversification or by increasing reuse of stereotyped solutions. Extending the Dynamic Intelligence Ceiling framework to richer embodied or open-world environments remains an important direction for future work.

\subsection{Implications and Scope}

Dynamic Intelligence Ceilings do not imply unbounded intelligence. All artificial systems remain constrained by physical resources, time, and informational access. The contribution of this work lies in clarifying that the most consequential limits are not absolute ceilings, but the mechanisms by which ceilings become prematurely fixed. While our benchmark focuses on planning and design, the DIC framework is domain-agnostic and may be instantiated in other contexts, including embodied interaction, scientific discovery, or multi-agent coordination.

The present study is methodological in nature. Our objective is not to compare specific AI systems, but to provide a diagnostic framework for identifying frontier saturation and drift across developmental trajectories. Applying the proposed metrics to concrete agents—such as reinforcement learning or language-based systems—is a natural next step and is the subject of ongoing and future work.

Because PDC and CDR are defined independently of task semantics, the proposed metrics can, in principle, be instantiated in domains such as robotics, scientific discovery, or multi-agent systems, provided that difficulty can be parametrically controlled.

\section{Conclusion}
\label{sec:conclusion}

We have introduced the concept of a Dynamic Intelligence Ceiling and demonstrated how it can be empirically estimated using trajectory-centric metrics and a procedurally generated benchmark. By shifting the focus from static performance to the evolution of performance frontiers, this work provides a principled framework for diagnosing premature convergence in artificial systems.

Rather than seeking unbounded intelligence, we argue for systems capable of sustaining development by delaying or preventing early fixation of their limits. Dynamic Intelligence Ceilings offer a concrete lens through which such systems can be evaluated and compared. We view this framework as a step toward a science of intelligence concerned not only with what artificial systems can achieve at a given moment but with how long—and in what manner—they can continue to grow.

\bibliographystyle{plainnat}

\end{document}